\documentclass[sigconf, authorversion, nonacm]{acmart}
\usepackage[font=small,labelfont=bf]{caption}
\usepackage{multirow}
\usepackage{array}
\usepackage{pgfplots}
\usepackage{subcaption} 
\usepackage{tikz}
\usepackage{pgfplots}

\AtBeginDocument{%
  \providecommand\BibTeX{{%
    \normalfont B\kern-0.5em{\scshape i\kern-0.25em b}\kern-0.8em\TeX}}}
\usepackage[normalem]{ulem}
\useunder{\uline}{\ul}{}
\copyrightyear{2025}
\acmYear{2025}

\newcolumntype{L}{>{\centering\arraybackslash}m{1.1cm}}
\begin{document}
\title{Optimization Strategies for Enhancing
Resource Efficiency in Transformers \&
Large Language Models\\}

\author{Tom Wallace}
\email{tw18dw@brocku.ca}
\affiliation{%
  \institution{Brock University}
  \city{St. Catharines}
  \state{Ontario}
  \country{Canada}
}
\author{Beatrice Ombuki-Berman}
\email{bombuki@brocku.ca}
\affiliation{%
  \institution{Brock University}
  \city{St. Catharines}
  \state{Ontario}
  \country{Canada}
}
\author{Naser Ezzati-Jivan}
\email{nezzati@brocku.ca}
\affiliation{%
  \institution{Brock University}
  \city{St. Catharines}
  \state{Ontario}
  \country{Canada}
}

\begin{abstract}
Advancements in Natural Language Processing are heavily reliant on the Transformer architecture, whose improvements come at substantial resource costs due to ever-growing model sizes. This study explores optimization techniques, including Quantization, Knowledge Distillation, and Pruning, focusing on energy and computational efficiency while retaining performance. Among standalone methods, 4-bit Quantization significantly reduces energy use with minimal accuracy loss. Hybrid approaches, like NVIDIA's Minitron approach combining KD and Structured Pruning, further demonstrate promising trade-offs between size reduction and accuracy retention. A novel optimization equation is introduced, offering a flexible framework for comparing various methods. Through the investigation of these compression methods, we provide valuable insights for developing more sustainable and efficient LLMs, shining a light on the often-ignored concern of energy efficiency.\footnote{A full list of links to all models used, in addition to supplemental data, is provided at https://github.com/OptimizationStrategies/Optimization-Strategies.}
\end{abstract}

\maketitle

\vspace{-3mm}
\section{Introduction}
While the Transformer architecture \cite{vaswani2023attention} achieves notable flexibility and precision across a wide range of language tasks, its performance comes at a high computational cost. The self-attention mechanism in large-scale models like OpenAI's GPT series requires significant processing power and memory to both train and operate, leading to substantial costs; At the launch of GPT-4, OpenAI’s CEO Sam Altman shared that the training process took "more than [\$100 Million]" \cite{Knight_2023} in total costs, factoring in salaries, energy, and hardware infrastructure.

Existing concerns over energy usage and environmental impact are intensified by the trend of continuously expanding model sizes and datasets, which further drive up energy consumption and financial expenses. Analysts suggest GPT-4 has upwards of 1 trillion parameters, dwarfing GPT-3’s 175 billion. As data centers frequently rely on water cooling, water consumption arises as an additional concern; Li et al. \cite{li2023making} estimate that the process of training GPT-3 directly used 700,000 liters of freshwater, with the model additionally needing to "‘drink’ a 500ml bottle of water for a simple conversation of roughly 20-50 questions and answers" \cite{li2023making}. This combination of economic, environmental, and resource concerns necessitates the formulation of innovative, systematic approaches to lowering costs while retaining as much performance as possible.

Methods such as pruning, quantization, and knowledge distillation have emerged as foundational methods for reducing both computational and financial overhead \cite{Hohman_2024}. These techniques aim to reduce model size, improve efficiency, and minimize resource consumption without sacrificing significant performance. However, each method presents its own set of challenges and trade-offs; the key lies in systematically comparing the effectiveness of these techniques, identifying scenarios where each approach excels, and exploring novel hybrid methods that combine the strengths of existing solutions. This paper aims to contribute to that goal by exploring several prominent compression methods, evaluating their effectiveness, and determining which may provide a more sustainable path forward for the development and deployment of future LLMs.

This paper makes the following contributions:
\begin{itemize}
    \item Presents a series of tests evaluating the changes in perplexity, energy usage, and computational speed of GPT-2 and OPT Transformer models when subjected to different optimization methods. Different combinations of Quantization, Knowledge Distillation, Attention Head Pruning, and Structured Pruning are all utilized where possible, with the goal of observing how the methods work in conjunction.
    \item Introduces an optimization equation which can be used to better evaluate which of the aforementioned methods are most suitable with regards to optimizing energy usage and/or computational speed.
    \item Applies said optimization equation to the results obtained from the method tests to create recommendations for which methods most effectively reduce costs, while minimizing perplexity loss.
    \item Performs an additional round of tests on a selection of state-of-the-art models created using either experimental methods or a novel combination of methods. These tests involve, in addition to the previous round's perplexity and time/energy tests, a set of knowledge \& comprehension benchmarks.
\end{itemize}
In summary, our primary goal is to evaluate the viability of various model compression methods in addressing critical trade-offs between time, energy consumption, and model accuracy. In particular, we aim to place a higher focus on energy efficiency; by highlighting the environmental impact of LLMs, often overlooked in preceding research, we hope to initiate a crucial dialogue on mitigating the energy demands of AI systems.

\section{Related Work}

This section reviews major advancements in model compression methods, focusing on quantization, knowledge distillation, and pruning. We also cover hybrid methods that combine these strategies to enhance efficiency and performance. This overview contextualizes our approach by identifying key contributions that inform our experimental design.
\subsection{Fundamental Methods}
Several methods exist which are relatively simple to implement into a model, and are often used as a basis for further experimentation and progress. 
\begin{itemize}
    \item \textbf{Quantization:} The conversion of a model's parameters to a smaller \& more efficient format, typically 8 or 4-bit integers \cite{dettmers2023spqr} \cite{dettmers2023qlora} \cite{dettmers2023case} \cite{dettmers2022llm} \cite{DBLP:journals/corr/abs-2110-02861}.

    \item \textbf{Knowledge Distillation:} The training of a smaller, faster "student" model to emulate a larger "teacher" model’s performance \cite{sanh2020distilbert}. By targeting the 'soft label' predictions from the teacher, the student model can closer approach the output of the teacher with a reduced risk of overfitting.

    \item \textbf{Structured Pruning:} The reduction of model size by removing a standardized proportion of less important parameters, minimizing the computational load \cite{HAGIWARA1994207, frantar2023sparsegpt}.

    \item \textbf{Attention Head Pruning:} A specialized form of pruning where Attention heads, the module within Transformer models which learns linguistic connections, are removed based on the quality of said connections \cite{voita-etal-2019-analyzing, michel2019sixteen}.
\end{itemize}

\subsection{Advanced Hybrid Methods}
Research has increasingly focused on hybrid methods that combine KD, pruning, and various other methods, often employing improved training approaches which aim to retain a model's knowledge. Additionally, research is constantly performed into method alternatives and modifications, which aim to investigate whether the existing industry standards may be improved.

\begin{itemize}
    \item \textbf{MiniLLM: A Refined Distillation Approach:} MiniLLM \cite{gu2024minillmknowledgedistillationlarge} refines KD by using reverse Kullback-Leibler divergence (KLD) to better align the student with the teacher’s high-confidence predictions.

    \item \textbf{LLM Pruning and Distillation in Practice: The Minitron Approach:} NVIDIA’s compact model approach \cite{muralidharan2024compactlanguagemodelspruning} combines structured pruning and KD, alongside accuracy retention techniques. Their results show substantial efficiency gains with minimal retraining and accuracy loss.

    \item \textbf{Sheared LLaMA: Accelerating Language Model Pre-training via Structured Pruning:} Sheared LLAMA \cite{xia2024shearedllamaacceleratinglanguage} introduces Targeted Structured Pruning, optimizing LLMs to predefined sizes by selectively pruning layers, attention heads, and dimensions based on model structure. 

\begin{figure}[h]
    \centering
    \includegraphics[width=0.8\columnwidth]{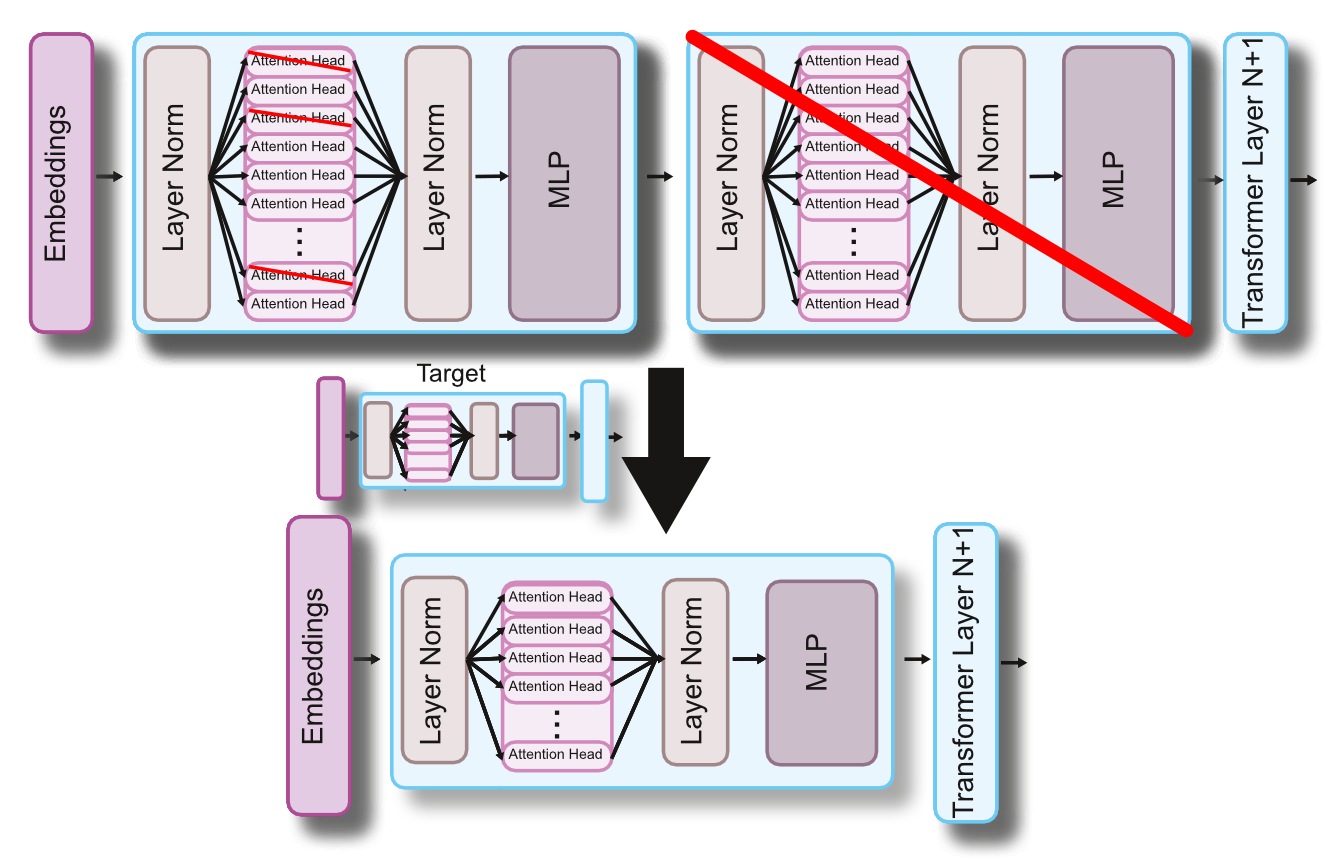}
    \caption{Targeted Structured Pruning: layers are pruned to match specific architecture sizes, ensuring efficiency with minimal retraining.}
    \label{fig:structured_pruning}
    \end{figure}
\end{itemize}

The reviewed methods demonstrate significant progress in model compression, each offering unique advantages. Quantization has been successful in reducing memory footprint, while knowledge distillation enables efficient student models that mimic larger teachers with minimal performance loss. Pruning methods, particularly structured approaches, help reduce computational demands without sacrificing core model structure, and the advanced techniques illustrate the benefits of combining these strategies or searching for potential improvements upon tried-and-true methods. 

However, while each technique offers clear advantages, existing studies often evaluate them in isolation or within limited contexts. To bridge this gap, our study systematically assesses these compression methods across varying model configurations and performance metrics, applying both individual and combined approaches. We introduce an optimization framework that quantifies the trade-offs between computational efficiency and accuracy retention, providing a robust basis for determining optimal compression strategies for deployment-specific requirements.

In the following sections, we present our methodology for testing these techniques on GPT-2 and OPT models, our optimization equation, and an in-depth analysis of experimental results that inform our recommendations for resource-efficient large language model deployments.

\section{Methodology}

We propose a systematic approach to identify and implement optimization techniques that achieve meaningful reductions in energy consumption and computational costs without sacrificing core model performance. Our hypothesis is that by carefully selecting and combining modular and advanced compression methods, LLMs can be adapted to run effectively in resource-constrained environments, broadening access to these powerful tools while promoting more sustainable AI practices.

Our methodology tests a range of optimization strategies across two levels of models—basic and advanced. Basic models are optimized with accessible, modular techniques that can be readily applied and adapted by users with standard computing resources. Advanced models incorporate cutting-edge compression methods developed by industry leaders, leveraging sophisticated combinations of techniques to push the boundaries of resource efficiency. This dual-level approach not only showcases practical optimizations for common use cases but also provides insights into the potential of high-performance methods for industry-grade applications.

To rigorously assess these techniques, we introduce a custom optimization equation that balances trade-offs between perplexity, energy usage, and computational time. This equation allows us to evaluate each technique’s impact on resource savings and performance retention, tailored to various priority settings (e.g., time vs. energy). By establishing a quantitative framework, our methodology guides researchers and practitioners in selecting the optimal configuration for their specific needs, from individual developers to large-scale deployments.

\subsection{Optimization Techniques}

This study employs four foundational optimization techniques selected to reduce computational costs and resource demands while preserving performance. Each technique offers unique benefits and has been chosen to maximize specific aspects of efficiency.

\begin{itemize}
    \item \textbf{Quantization}: Quantization reduces the precision of model weights, typically from 32-bit floating points to lower precisions like 8-bit or 4-bit. This approach significantly decreases memory usage and energy consumption, simplifying the computations performed during training and inference with only minimal loss of model accuracy. The technique is broadly applied in situations where memory usage is a major concern, such as deployment on lower-memory devices like laptops or phones. We leverage \emph{Huggingface} and the \emph{BitsandBytes} library \cite{dettmers2023spqr, dettmers2023qlora, dettmers2023case, dettmers2022llm, DBLP:journals/corr/abs-2110-02861} for efficient runtime conversion to 8/4-bit formats with minimal accuracy degradation.

    \item \textbf{Knowledge Distillation}: Knowledge distillation involves training a smaller “student” model to replicate the outputs of a larger, high-performing “teacher” model \cite{sanh2020distilbert}. This technique was chosen for its ability to reduce model size while retaining much of the original model’s knowledge and performance. It is especially useful for scenarios that require compact, efficient models without substantial accuracy degradation.

    \item \textbf{Attention Head Pruning}: In Transformer models, multiple attention heads are responsible for processing different parts of the input. Attention head pruning removes the less essential heads \cite{voita-etal-2019-analyzing, michel2019sixteen}, which decreases model complexity and resource use. This technique allows reductions in resource requirements while preserving the core structure of the model. However, it is less effective in generative models like GPT than in encoder-based models, which limits its use cases in some applications. The methodology we utilize for choosing which attention heads to remove is based on the model's attention output given an input sentence. This output contains the attention values given by each attention head; If a single token receives most of the attention, the head likely does not contain any important connections and can safely be pruned. We perform separate experiments using two different thresholds for sake of comparison; The first threshold is a single token receiving over 90\% of attention, and the second threshold is 80\%.

    \item \textbf{Magnitude Pruning}: This method removes model parameters with the smallest absolute values, which tend to have lower impact on overall performance. However, most forms of pruning often require retraining to maintain accuracy, making it a resource-intensive method to implement. Pruning methods for LLMs like SparseGPT \cite{frantar2023sparsegpt} aim for defined sparsity structures to maintain accuracy while improving efficiency. SparseGPT’s 2:4 structured sparsity, which we employ, achieves practical application benefits while retaining model architecture. 
\end{itemize}

Each technique is applied both independently and in various combinations on basic models. This approach allows us to assess their effectiveness in enhancing energy efficiency, speed, and model accuracy. Table \ref{tab:my_label} lists the different combinations of techniques tested in our experiments, giving an overview of each method's distinct setup.

\begin{table}[h]
    \centering
    \begin{tabular}{|l|}
\hline
Knowledge Distilled, 90\% AH        \\ \hline
Knowledge Distilled, 80\% AH        \\ \hline
Knowledge Distilled, 8-Bit          \\ \hline
Knowledge Distilled, 4-Bit          \\ \hline
8-Bit, 90\% AH                      \\ \hline
8-Bit, 80\% AH                      \\ \hline
4-Bit, 90\% AH                      \\ \hline
4-Bit, 80\% AH                      \\ \hline
Knowledge Distilled, 8-Bit, 90\% AH \\ \hline
Knowledge Distilled, 8-Bit, 80\% AH \\ \hline
Knowledge Distilled, 4-Bit, 90\% AH \\ \hline
Knowledge Distilled, 4-Bit, 80\% AH \\ \hline
\end{tabular}
    \caption{List of all experiment combinations performed.}
    \label{tab:my_label}
\end{table}

\subsection{Basic Models and Application of Techniques}

We use a set of basic models—versions of GPT-2 and OPT—as the foundation for evaluating these techniques. These models are accessible, widely used in research, and enable a modular approach to testing optimizations. Our experiments aim to observe the effect of each technique when applied to standard models in different configurations and model sizes, allowing for a practical assessment of each technique’s effectiveness.

\begin{itemize}
    \item \textbf{Models Tested}: GPT-2 and OPT models at different parameter scales, including standard GPT-2 (125M parameters), GPT-2 Large (774M parameters), and GPT-2 XL (1.5B parameters). Meta's similarly-sized OPT models are also tested as comparisons where applicable.
   
    \item \textbf{Experiment Setup}: Each model is subjected to the optimization techniques, both individually and in various combinations (e.g., Quantization with Attention Head Pruning). Performance metrics such as perplexity, energy usage, and computation time are recorded. The experiments help identify the effectiveness of each technique and combination, especially for applications on consumer-grade hardware.

    \item \textbf{Goal}: This phase provides insight into user-friendly optimizations that can be applied by developers without extensive computational resources, focusing on standalone methods that are modular and accessible.
\end{itemize}

\subsection{Advanced Models and Hybrid Techniques}

To evaluate high-performance optimization strategies, we test advanced, pre-compressed models created by industry researchers. These models, including MiniLLM \cite{gu2024minillmknowledgedistillationlarge}, MN-Minitron \cite{muralidharan2024compactlanguagemodelspruning}, and ShearedLlama \cite{xia2024shearedllamaacceleratinglanguage}, apply hybrid or modified optimization techniques to achieve significant compression without a substantial drop in performance. Table \ref{tab:models2} lists all advanced models tested, showing their respective compression techniques and parameter counts. To provide a reference point for how these methods affect the models' performance and efficiency, we test these models' progenitors; the pre-pruning models in the case of pruning methods, and the teacher models in the case of Knowledge Distillation.

\begin{itemize}
    \item \textbf{Advanced Models Tested}: 
        \begin{itemize}
            \item \textbf{MiniLLM}: MiniLLM \cite{gu2024minillmknowledgedistillationlarge} optimizes the standard KD approach by using reverse Kullback-Leibler divergence (KLD); In contrast to standard KLD, which averages across the entire probability field, Reverse KLD focuses solely on the highest probablility output. By changing the training priorities, MiniLLM achieves superior performance compared to standard KD in various LLM architectures across instruction-following tasks. This paper uses MiniLLM-trained OPT and Llama models, alongside models trained using Sequenced KD (SeqKD), standard KD, and from-scratch training, for comparative purposes.
            \item \textbf{Minitron (NVIDIA)}: NVIDIA’s compact model approach \cite{muralidharan2024compactlanguagemodelspruning, sreenivas2024llmpruningdistillationpractice} combines structured pruning and KD; A large, high-quality model is pruned, with the pruned version trained by the unpruned version. Models trained using this approach retained a majority of accuracy on benchmarks like Winogrande and MMLU, proving that model sizes can be heavily reduced without notable performance loss. For example, MN-Minitron reduces model size by a third with minimal accuracy loss on most tasks, and minor accuracy gains on some \cite{muralidharan2024compactlanguagemodelspruning}.
            \item \textbf{ShearedLlama}: Uses a novel technique, targeted structured pruning, to optimize LLMs. This method reduces models to predefined sizes by selectively pruning layers, attention heads, and dimensions based on predefined model structures \cite{xia2024shearedllamaacceleratinglanguage}. This method significantly reduces computational requirements and necessary training time compared to conventional methods. 
        \end{itemize}
    \item \textbf{Techniques in Advanced Models}: These advanced models either apply a combination of techniques—such as structured pruning with knowledge distillation—or an experimental modification of existing techniques; While challenging to implement using typical hardware, they often yield substantial reductions in both model size and resource consumption.

    \item \textbf{Goal}: Testing these models helps benchmark the effectiveness of hybrid techniques, offering insights into highly efficient model designs for resource-constrained environments.
\end{itemize}

\begin{table}[h]
\footnotesize
    \centering
\begin{tabular}{l|ll}
Baseline Reference    &                         &                     \\ \hline
GPT2-XL (1.5B)     & MiniLLM GPT2XL (774M)   & Base GPT2L (774M)   \\ 
                & KD GPT2XL (774M)      & SeqKD GPT2XL (774M)     \\ \hline
OPT-13B            & MiniLLM OPT (6.7B)      & Base OPT (6.7B)     \\
                    & KD OPT (6.7B)           & SeqKD OPT (6.7B)    \\
Llama (13B)        & MiniLLM Llama (6.7B)    & Base Llama (6.7B)   \\  
                   & KD Llama (6.7B)         & SeqKD Llama (6.7B)  \\ \hline
Llama2 (7B)        & ShearedLlama (2.7B)     & ShearedLlama (1.3B) \\ \hline
Mistral-Nemo (12B) & MN-Minitron (8B)        &                     \\ \hline
Llama-3.1 (8B)     & Llama-3.1-Minitron (4B) &                     \\ \hline
\end{tabular}
    \caption{List of All Models Tested}
    \label{tab:models2}
\end{table}

\subsection{Optimization Equation}
To objectively evaluate and compare the performance-resource trade-offs for each optimization method and model, we developed an optimization equation, allowing us to compare these optimization methods by balancing the resulting perplexity increase against other critical factors: energy usage and computational time. This equation incorporates adjustable weights to prioritize either energy efficiency, computational speed, or a balanced approach, depending on the user’s needs.

The equation, defined as:
\begin{equation}
    opt = P_c^{1.5}(\alpha T_c + \beta E_c)
\end{equation}
where \( P_c \), \( T_c \), and \( E_c \) represent the change ratios of perplexity, time, and energy, respectively, between the optimized and base models. The weights \( \alpha \) and \( \beta \) may be adjusted to align with specific priorities—favoring energy savings, speed, or an even balance of both. In order to ensure perplexity doesn’t increase to the point of making the model worthless, its factor is raised to the power of 1.5, penalizing high perplexity increases more than efficiency benefits are rewarded. 

The output of this equation, \emph{opt}, is a combined factor of the relative changes in perplexity, energy usage, and time; A lower output value is more desirable, indicating a lower increase in perplexity with respect to the decrease in costs. Three sample configurations we apply in our experiments are listed in Table \ref{tab:alphabeta}. By applying this equation in each experimental setup, we provide a structured method to assess whether any decrease in perplexity (i.e., performance drop) is justified by the achieved savings in energy and computational time, facilitating informed decisions on optimization strategy.
\subsubsection{Non-Perplexity \& Additional Measurements}
It is important to note that, when measuring perplexity, lower values are more desirable; When utilizing a scoring method where higher values are preferred, one should adjust the equation for \(P_c\) to instead be \(P_c=\frac{P_{m}}{P_{b}}\).

Additionally, some testing methodologies may utilize several tests across various knowledge benchmarks; When doing so, we adjust our calculation of \(P_c\) as such:
\begin{itemize}
    \item For \emph{i} measurements considered:
    \item \(P_{cn}=\frac{P_{bn}}{P_{mn}}\) for an \emph{n}th measurement where lower values are desired (e.g. Perplexity),
    \item \(P_{cn}=\frac{P_{mn}-x\frac{4}{5}}{P_{bn}-x\frac{4}{5}}\) for an \emph{n}th measurement where higher values are desired, with an expected minimum score of x (e.g. The expected score achieved by selecting answers at random). x is multiplied by 4/5 to give a small safety net to models which perform worse than random.
    \item \(P_c = (\sum_{n=1}^{\infty}P_{cn}^{1.5}\))/i.
\end{itemize}
In summary, we find the average of all \(P_c\) values across each measurement, using that as \(P_c\) in our base calculation. By raising each \(P_c\) to the power of 1.5 before averaging the values, higher drops in any particular score are more significantly penalized.

\begin{table}[]
\begin{tabular}{|l|l|l|}
\hline
Parameter Sets & Alpha (Time Weight) & Beta (Energy Weight) \\ \hline
Balanced       & 0.5                 & 0.5                  \\ \hline
Energy Focus   & 0.1                 & 0.9                  \\ \hline
Runtime Focus  & 0.9                 & 0.1                  \\ \hline
\end{tabular}
\caption{Parameter Sets for Optimization Equation Weights}
\label{tab:alphabeta}
\end{table}

\section{Evaluation and Results}

This section presents an analysis of the impact of various optimization techniques on large language models (LLMs), focusing on achieving a balance between performance retention and resource efficiency. 
\subsection{Experimental Setup}

The experiments were conducted on a system equipped with an NVIDIA GeForce 4070TI Graphics Card (12GB VRAM) for GPU-compatible methods and a 2.1GHz Intel Platinum 8160F Skylake processor for techniques like Magnitude Pruning, which is not GPU-compatible. The metrics chosen for analysis—perplexity, computational time, and energy consumption—directly represent the primary concerns of most users. These metrics are essential for evaluating the applicability of optimized LLMs in various deployment contexts, from industry-scale service deployments to mobile phone applications.

For each model and testing dataset pair, the perplexity evaluation algorithm is run 30 times to reduce the impact of random fluctuations. The Pruning method is an exception; due to the significantly reduced speed of CPU-based inference, it runs only 5 times. Using the Carbontracker\cite{anthony2020carbontracker} library, we measure total energy usage, time taken, and estimated carbon output across the 30 runs.

\subsection{Evaluation of Standalone Techniques on Basic Models}
To establish a baseline, we first analyze the impact of standalone optimization techniques individually applied to basic models. The table below presents the results across different scales of the GPT-2 model series, comparing changes in perplexity, runtime, and energy consumption.

\begin{table*}[h!]
\centering
\footnotesize
\begin{tabular}{|l|c|c|c|c|c|c|}
\hline
\textbf{Technique}       & \textbf{Perplexity} & \textbf{Perplexity Change (\%)} & \textbf{Runtime (s)} & \textbf{Runtime Change (\%)} & \textbf{Energy (kWh)} & \textbf{Energy Change (\%)} \\ \hline
Base                      & 34.29               & -                               & 212.33               & -                             & 0.02335               & -                            \\ \hline
8-bit Quantization        & 34.40               & +0.31\%                         & 644.67               & +203.78\%                     & 0.01017               & -56.28\%                     \\ \hline
4-bit Quantization       & 35.56               & +3.79\%                         & 222.67               & +4.86\%                       & 0.01162               & -50.21\%                     \\ \hline
Knowledge Distillation   & 49.41               & +44.54\%                        & 131.67               & -37.91\%                      & 0.01395               & -40.30\%                     \\ \hline
AH90                   & 45.60203   & +34.01\%                   & 203.6667    & -4.10\%                & 0.022225     & -4.67\%               \\ \hline
AH80                   & 52.24299   & +53.94\%                   & 197.6667    & -6.99\%                & 0.021458     & -8.04\%               \\ \hline
Base (OPT-125M)              & 34.713     &                          & 5497.333    &                       & 0.36800      &                      \\ \hline
Pruning (OPT-125M)           & 49.512     & +43.74\%                   & 4650        & -16.49\%               & 0.27459      & -21.78\%              \\ \hline
\end{tabular}
\caption{Results for Standalone Techniques on GPT-2 (125M)}
\label{tab:gpt2_results}
\end{table*}

Table \ref{tab:gpt2_results} shows that standalone techniques exhibit varied trade-offs between efficiency and accuracy. The two Quantization methods achieve high energy savings with negligible increases in perplexity, albeit at the cost of significantly higher processing times in the case of 8-bit, and a minor increase in runtime in the case of 4-bit. Meanwhile, Knowledge Distillation reduces runtime and energy usage but at a notable cost in accuracy. Both forms of the Attention Head Pruning method provide small decreases in time and energy usage, vastly outweighed by a significant perplexity increase. The SparseGPT pruning method results in a similar perplexity increase as Knowledge Distillation, but with less significant decreases in time and energy. This indicates that standalone techniques can be effective in optimizing specific aspects of model performance but may not be ideal for applications requiring balanced improvements across all metrics.

\subsection{Analysis of Resource-Performance Trade-offs}
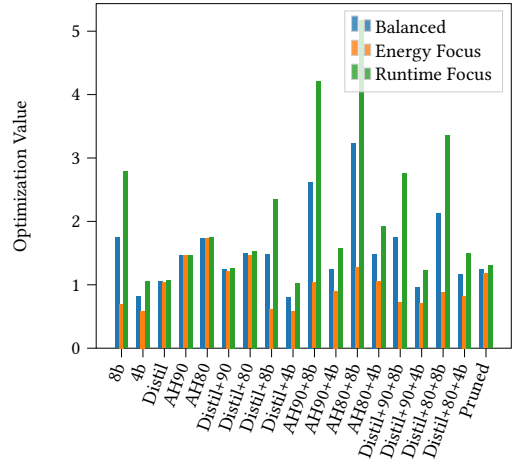
\begin{figure}
    \centering
    \resizebox{0.8\columnwidth}{!}{
\begin{tikzpicture}

\definecolor{darkgray176}{RGB}{176,176,176}
\definecolor{darkorange25512714}{RGB}{255,127,14}
\definecolor{forestgreen4416044}{RGB}{44,160,44}
\definecolor{lightgray204}{RGB}{204,204,204}
\definecolor{steelblue31119180}{RGB}{31,119,180}

\begin{axis}[
legend cell align={left},
legend style={fill opacity=0.8, draw opacity=1, text opacity=1, draw=lightgray204},
tick align=outside,
tick pos=left,
x grid style={darkgray176},
xmin=-1.18, xmax=18.18,
xtick style={color=black},
xtick={0,1,2,3,4,5,6,7,8,9,10,11,12,13,14,15,16,17},
xticklabel style={rotate=70.0,anchor=east},
xticklabels={
  8b,
  4b,
  Distil,
  AH90,
  AH80,
  Distil+90,
  Distil+80,
  Distil+8b,
  Distil+4b,
  AH90+8b,
  AH90+4b,
  AH80+8b,
  AH80+4b,
  Distil+90+8b,
  Distil+90+4b,
  Distil+80+8b,
  Distil+80+4b,
  Pruned
},
y grid style={darkgray176},
ylabel={Optimization Value},
ymin=0, ymax=5.434695,
ytick style={color=black}
]
\draw[draw=none,fill=steelblue31119180] (axis cs:-0.3,0) rectangle (axis cs:-0.1,1.7439);
\addlegendimage{ybar,ybar legend,draw=none,fill=steelblue31119180}
\addlegendentry{Balanced}

\draw[draw=none,fill=steelblue31119180] (axis cs:0.7,0) rectangle (axis cs:0.9,0.8165);
\draw[draw=none,fill=steelblue31119180] (axis cs:1.7,0) rectangle (axis cs:1.9,1.0526);
\draw[draw=none,fill=steelblue31119180] (axis cs:2.7,0) rectangle (axis cs:2.9,1.4651);
\draw[draw=none,fill=steelblue31119180] (axis cs:3.7,0) rectangle (axis cs:3.9,1.7391);
\draw[draw=none,fill=steelblue31119180] (axis cs:4.7,0) rectangle (axis cs:4.9,1.2413);
\draw[draw=none,fill=steelblue31119180] (axis cs:5.7,0) rectangle (axis cs:5.9,1.4976);
\draw[draw=none,fill=steelblue31119180] (axis cs:6.7,0) rectangle (axis cs:6.9,1.482);
\draw[draw=none,fill=steelblue31119180] (axis cs:7.7,0) rectangle (axis cs:7.9,0.8068);
\draw[draw=none,fill=steelblue31119180] (axis cs:8.7,0) rectangle (axis cs:8.9,2.6221);
\draw[draw=none,fill=steelblue31119180] (axis cs:9.7,0) rectangle (axis cs:9.9,1.2372);
\draw[draw=none,fill=steelblue31119180] (axis cs:10.7,0) rectangle (axis cs:10.9,3.2243);
\draw[draw=none,fill=steelblue31119180] (axis cs:11.7,0) rectangle (axis cs:11.9,1.4838);
\draw[draw=none,fill=steelblue31119180] (axis cs:12.7,0) rectangle (axis cs:12.9,1.7456);
\draw[draw=none,fill=steelblue31119180] (axis cs:13.7,0) rectangle (axis cs:13.9,0.9654);
\draw[draw=none,fill=steelblue31119180] (axis cs:14.7,0) rectangle (axis cs:14.9,2.1206);
\draw[draw=none,fill=steelblue31119180] (axis cs:15.7,0) rectangle (axis cs:15.9,1.1634);
\draw[draw=none,fill=steelblue31119180] (axis cs:16.7,0) rectangle (axis cs:16.9,1.2408);
\draw[draw=none,fill=darkorange25512714] (axis cs:-0.1,0) rectangle (axis cs:0.1,0.6987);
\addlegendimage{ybar,ybar legend,draw=none,fill=darkorange25512714}
\addlegendentry{Energy Focus}

\draw[draw=none,fill=darkorange25512714] (axis cs:0.9,0) rectangle (axis cs:1.1,0.5837);
\draw[draw=none,fill=darkorange25512714] (axis cs:1.9,0) rectangle (axis cs:2.1,1.0368);
\draw[draw=none,fill=darkorange25512714] (axis cs:2.9,0) rectangle (axis cs:3.1,1.4605);
\draw[draw=none,fill=darkorange25512714] (axis cs:3.9,0) rectangle (axis cs:4.1,1.73);
\draw[draw=none,fill=darkorange25512714] (axis cs:4.9,0) rectangle (axis cs:5.1,1.2166);
\draw[draw=none,fill=darkorange25512714] (axis cs:5.9,0) rectangle (axis cs:6.1,1.465);
\draw[draw=none,fill=darkorange25512714] (axis cs:6.9,0) rectangle (axis cs:7.1,0.6212);
\draw[draw=none,fill=darkorange25512714] (axis cs:7.9,0) rectangle (axis cs:8.1,0.5873);
\draw[draw=none,fill=darkorange25512714] (axis cs:8.9,0) rectangle (axis cs:9.1,1.0399);
\draw[draw=none,fill=darkorange25512714] (axis cs:9.9,0) rectangle (axis cs:10.1,0.8997);
\draw[draw=none,fill=darkorange25512714] (axis cs:10.9,0) rectangle (axis cs:11.1,1.2726);
\draw[draw=none,fill=darkorange25512714] (axis cs:11.9,0) rectangle (axis cs:12.1,1.0505);
\draw[draw=none,fill=darkorange25512714] (axis cs:12.9,0) rectangle (axis cs:13.1,0.7295);
\draw[draw=none,fill=darkorange25512714] (axis cs:13.9,0) rectangle (axis cs:14.1,0.7021);
\draw[draw=none,fill=darkorange25512714] (axis cs:14.9,0) rectangle (axis cs:15.1,0.8763);
\draw[draw=none,fill=darkorange25512714] (axis cs:15.9,0) rectangle (axis cs:16.1,0.8239);
\draw[draw=none,fill=darkorange25512714] (axis cs:16.9,0) rectangle (axis cs:17.1,1.1786);
\draw[draw=none,fill=forestgreen4416044] (axis cs:0.1,0) rectangle (axis cs:0.3,2.7891);
\addlegendimage{ybar,ybar legend,draw=none,fill=forestgreen4416044}
\addlegendentry{Runtime Focus}

\draw[draw=none,fill=forestgreen4416044] (axis cs:1.1,0) rectangle (axis cs:1.3,1.0492);
\draw[draw=none,fill=forestgreen4416044] (axis cs:2.1,0) rectangle (axis cs:2.3,1.0684);
\draw[draw=none,fill=forestgreen4416044] (axis cs:3.1,0) rectangle (axis cs:3.3,1.4697);
\draw[draw=none,fill=forestgreen4416044] (axis cs:4.1,0) rectangle (axis cs:4.3,1.7481);
\draw[draw=none,fill=forestgreen4416044] (axis cs:5.1,0) rectangle (axis cs:5.3,1.2659);
\draw[draw=none,fill=forestgreen4416044] (axis cs:6.1,0) rectangle (axis cs:6.3,1.5302);
\draw[draw=none,fill=forestgreen4416044] (axis cs:7.1,0) rectangle (axis cs:7.3,2.3429);
\draw[draw=none,fill=forestgreen4416044] (axis cs:8.1,0) rectangle (axis cs:8.3,1.0264);
\draw[draw=none,fill=forestgreen4416044] (axis cs:9.1,0) rectangle (axis cs:9.3,4.2044);
\draw[draw=none,fill=forestgreen4416044] (axis cs:10.1,0) rectangle (axis cs:10.3,1.5746);
\draw[draw=none,fill=forestgreen4416044] (axis cs:11.1,0) rectangle (axis cs:11.3,5.1759);
\draw[draw=none,fill=forestgreen4416044] (axis cs:12.1,0) rectangle (axis cs:12.3,1.9171);
\draw[draw=none,fill=forestgreen4416044] (axis cs:13.1,0) rectangle (axis cs:13.3,2.7618);
\draw[draw=none,fill=forestgreen4416044] (axis cs:14.1,0) rectangle (axis cs:14.3,1.2287);
\draw[draw=none,fill=forestgreen4416044] (axis cs:15.1,0) rectangle (axis cs:15.3,3.3649);
\draw[draw=none,fill=forestgreen4416044] (axis cs:16.1,0) rectangle (axis cs:16.3,1.5029);
\draw[draw=none,fill=forestgreen4416044] (axis cs:17.1,0) rectangle (axis cs:17.3,1.3029);
\end{axis}

\end{tikzpicture}}
    \caption{GPT-2 methods}
    \label{fig:gpt2opt}
\end{figure}

\begin{figure}
    \centering
    \resizebox{0.8\columnwidth}{!}{
\begin{tikzpicture}

\definecolor{darkgray176}{RGB}{176,176,176}
\definecolor{darkorange25512714}{RGB}{255,127,14}
\definecolor{forestgreen4416044}{RGB}{44,160,44}
\definecolor{lightgray204}{RGB}{204,204,204}
\definecolor{steelblue31119180}{RGB}{31,119,180}

\begin{axis}[
legend cell align={left},
legend style={
  fill opacity=0.8,
  draw opacity=1,
  text opacity=1,
  at={(0.03,0.97)},
  anchor=north west,
  draw=lightgray204
},
tick align=outside,
tick pos=left,
x grid style={darkgray176},
xmin=-1.13, xmax=17.13,
xtick style={color=black},
xtick={0,1,2,3,4,5,6,7,8,9,10,11,12,13,14,15,16},
xticklabel style={rotate=70.0,anchor=east},
xticklabels={
  8b,
  4b,
  Distil,
  AH90,
  AH80,
  Distil+90,
  Distil+80,
  Distil+8b,
  Distil+4b,
  AH90+8b,
  AH90+4b,
  AH80+8b,
  AH80+4b,
  Distil+90+8b,
  Distil+90+4b,
  Distil+80+8b,
  Distil+80+4b
},
y grid style={darkgray176},
ylabel={Optimization Value},
ymin=0, ymax=2.28382135065,
ytick style={color=black}
]
\draw[draw=none,fill=steelblue31119180] (axis cs:-0.3,0) rectangle (axis cs:-0.1,0.721524025);
\addlegendimage{ybar,ybar legend,draw=none,fill=steelblue31119180}
\addlegendentry{Balanced}

\draw[draw=none,fill=steelblue31119180] (axis cs:0.7,0) rectangle (axis cs:0.9,0.566166903);
\draw[draw=none,fill=steelblue31119180] (axis cs:1.7,0) rectangle (axis cs:1.9,0.558552933);
\draw[draw=none,fill=steelblue31119180] (axis cs:2.7,0) rectangle (axis cs:2.9,0.921462808);
\draw[draw=none,fill=steelblue31119180] (axis cs:3.7,0) rectangle (axis cs:3.9,1.007529558);
\draw[draw=none,fill=steelblue31119180] (axis cs:4.7,0) rectangle (axis cs:4.9,0.713729206);
\draw[draw=none,fill=steelblue31119180] (axis cs:5.7,0) rectangle (axis cs:5.9,0.861300871);
\draw[draw=none,fill=steelblue31119180] (axis cs:6.7,0) rectangle (axis cs:6.9,0.832722365);
\draw[draw=none,fill=steelblue31119180] (axis cs:7.7,0) rectangle (axis cs:7.9,0.634306725);
\draw[draw=none,fill=steelblue31119180] (axis cs:8.7,0) rectangle (axis cs:8.9,0.849355554);
\draw[draw=none,fill=steelblue31119180] (axis cs:9.7,0) rectangle (axis cs:9.9,0.56763242);
\draw[draw=none,fill=steelblue31119180] (axis cs:10.7,0) rectangle (axis cs:10.9,0.990876463);
\draw[draw=none,fill=steelblue31119180] (axis cs:11.7,0) rectangle (axis cs:11.9,0.63388384);
\draw[draw=none,fill=steelblue31119180] (axis cs:12.7,0) rectangle (axis cs:12.9,1.124782886);
\draw[draw=none,fill=steelblue31119180] (axis cs:13.7,0) rectangle (axis cs:13.9,0.786913198);
\draw[draw=none,fill=steelblue31119180] (axis cs:14.7,0) rectangle (axis cs:14.9,1.39907728);
\draw[draw=none,fill=steelblue31119180] (axis cs:15.7,0) rectangle (axis cs:15.9,1.004982471);
\draw[draw=none,fill=darkorange25512714] (axis cs:-0.1,0) rectangle (axis cs:0.1,0.371109658);
\addlegendimage{ybar,ybar legend,draw=none,fill=darkorange25512714}
\addlegendentry{Energy Focus}

\draw[draw=none,fill=darkorange25512714] (axis cs:0.9,0) rectangle (axis cs:1.1,0.531141216);
\draw[draw=none,fill=darkorange25512714] (axis cs:1.9,0) rectangle (axis cs:2.1,0.431650654);
\draw[draw=none,fill=darkorange25512714] (axis cs:2.9,0) rectangle (axis cs:3.1,0.967518508);
\draw[draw=none,fill=darkorange25512714] (axis cs:3.9,0) rectangle (axis cs:4.1,1.055418944);
\draw[draw=none,fill=darkorange25512714] (axis cs:4.9,0) rectangle (axis cs:5.1,0.554063969);
\draw[draw=none,fill=darkorange25512714] (axis cs:5.9,0) rectangle (axis cs:6.1,0.687412089);
\draw[draw=none,fill=darkorange25512714] (axis cs:6.9,0) rectangle (axis cs:7.1,0.398903894);
\draw[draw=none,fill=darkorange25512714] (axis cs:7.9,0) rectangle (axis cs:8.1,0.606355189);
\draw[draw=none,fill=darkorange25512714] (axis cs:8.9,0) rectangle (axis cs:9.1,0.394641462);
\draw[draw=none,fill=darkorange25512714] (axis cs:9.9,0) rectangle (axis cs:10.1,0.524166347);
\draw[draw=none,fill=darkorange25512714] (axis cs:10.9,0) rectangle (axis cs:11.1,0.448569404);
\draw[draw=none,fill=darkorange25512714] (axis cs:11.9,0) rectangle (axis cs:12.1,0.572364558);
\draw[draw=none,fill=darkorange25512714] (axis cs:12.9,0) rectangle (axis cs:13.1,0.508900814);
\draw[draw=none,fill=darkorange25512714] (axis cs:13.9,0) rectangle (axis cs:14.1,0.733040429);
\draw[draw=none,fill=darkorange25512714] (axis cs:14.9,0) rectangle (axis cs:15.1,0.623086607);
\draw[draw=none,fill=darkorange25512714] (axis cs:15.9,0) rectangle (axis cs:16.1,0.927757786);
\draw[draw=none,fill=forestgreen4416044] (axis cs:0.1,0) rectangle (axis cs:0.3,1.071938393);
\addlegendimage{ybar,ybar legend,draw=none,fill=forestgreen4416044}
\addlegendentry{Runtime Focus}

\draw[draw=none,fill=forestgreen4416044] (axis cs:1.1,0) rectangle (axis cs:1.3,0.60119259);
\draw[draw=none,fill=forestgreen4416044] (axis cs:2.1,0) rectangle (axis cs:2.3,0.685455212);
\draw[draw=none,fill=forestgreen4416044] (axis cs:3.1,0) rectangle (axis cs:3.3,0.875407108);
\draw[draw=none,fill=forestgreen4416044] (axis cs:4.1,0) rectangle (axis cs:4.3,0.959640172);
\draw[draw=none,fill=forestgreen4416044] (axis cs:5.1,0) rectangle (axis cs:5.3,0.873394442);
\draw[draw=none,fill=forestgreen4416044] (axis cs:6.1,0) rectangle (axis cs:6.3,1.035189654);
\draw[draw=none,fill=forestgreen4416044] (axis cs:7.1,0) rectangle (axis cs:7.3,1.266540837);
\draw[draw=none,fill=forestgreen4416044] (axis cs:8.1,0) rectangle (axis cs:8.3,0.662258262);
\draw[draw=none,fill=forestgreen4416044] (axis cs:9.1,0) rectangle (axis cs:9.3,1.304069647);
\draw[draw=none,fill=forestgreen4416044] (axis cs:10.1,0) rectangle (axis cs:10.3,0.611098494);
\draw[draw=none,fill=forestgreen4416044] (axis cs:11.1,0) rectangle (axis cs:11.3,1.533183522);
\draw[draw=none,fill=forestgreen4416044] (axis cs:12.1,0) rectangle (axis cs:12.3,0.695403121);
\draw[draw=none,fill=forestgreen4416044] (axis cs:13.1,0) rectangle (axis cs:13.3,1.740664958);
\draw[draw=none,fill=forestgreen4416044] (axis cs:14.1,0) rectangle (axis cs:14.3,0.840785967);
\draw[draw=none,fill=forestgreen4416044] (axis cs:15.1,0) rectangle (axis cs:15.3,2.175067953);
\draw[draw=none,fill=forestgreen4416044] (axis cs:16.1,0) rectangle (axis cs:16.3,1.082207156);
\end{axis}

\end{tikzpicture}}
    \caption{GPT2-XL methods}
    \label{fig:gpt2xloptv2}
\end{figure}
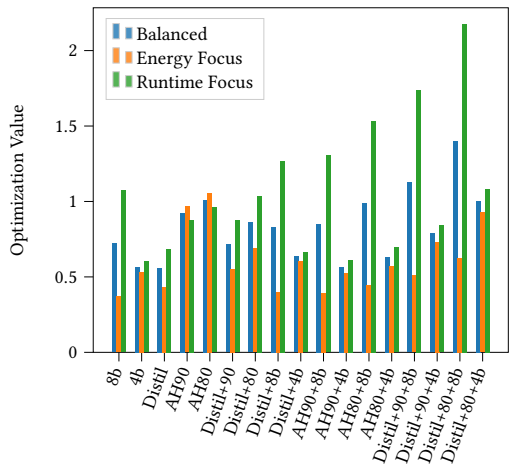

Using our custom optimization equation, we assess the trade-offs between perplexity, energy, and computational time. This equation allows adjusting weights on energy and computation time, guiding optimization toward specific deployment requirements. The trade-offs are summarized in figures ~\ref{fig:gpt2opt}, ~\ref{fig:gpt2largeopt}, and ~\ref{fig:gpt2xloptv2}.

\begin{figure}
    \centering
    \resizebox{0.8\columnwidth}{!}{
\begin{tikzpicture}

\definecolor{darkgray176}{RGB}{176,176,176}
\definecolor{darkorange25512714}{RGB}{255,127,14}
\definecolor{forestgreen4416044}{RGB}{44,160,44}
\definecolor{lightgray204}{RGB}{204,204,204}
\definecolor{steelblue31119180}{RGB}{31,119,180}

\begin{axis}[
legend cell align={left},
legend style={fill opacity=0.8, draw opacity=1, text opacity=1, draw=lightgray204},
tick align=outside,
tick pos=left,
x grid style={darkgray176},
xmin=-1.13, xmax=17.13,
xtick style={color=black},
xtick={0,1,2,3,4,5,6,7,8,9,10,11,12,13,14,15,16},
xticklabel style={rotate=70.0,anchor=east},
xticklabels={
  8b,
  4b,
  Distil,
  AH90,
  AH80,
  Distil+90,
  Distil+80,
  Distil+8b,
  Distil+4b,
  AH90+8b,
  AH90+4b,
  AH80+8b,
  AH80+4b,
  Distil+90+8b,
  Distil+90+4b,
  Distil+80+8b,
  Distil+80+4b
},
y grid style={darkgray176},
ylabel={Optimization Value},
ymin=0, ymax=2.675841,
ytick style={color=black}
]
\draw[draw=none,fill=steelblue31119180] (axis cs:-0.3,0) rectangle (axis cs:-0.1,1.07084);
\addlegendimage{ybar,ybar legend,draw=none,fill=steelblue31119180}
\addlegendentry{Balanced}

\draw[draw=none,fill=steelblue31119180] (axis cs:0.7,0) rectangle (axis cs:0.9,0.59813);
\draw[draw=none,fill=steelblue31119180] (axis cs:1.7,0) rectangle (axis cs:1.9,1.08424);
\draw[draw=none,fill=steelblue31119180] (axis cs:2.7,0) rectangle (axis cs:2.9,1.17478);
\draw[draw=none,fill=steelblue31119180] (axis cs:3.7,0) rectangle (axis cs:3.9,1.33334);
\draw[draw=none,fill=steelblue31119180] (axis cs:4.7,0) rectangle (axis cs:4.9,1.08101);
\draw[draw=none,fill=steelblue31119180] (axis cs:5.7,0) rectangle (axis cs:5.9,1.09981);
\draw[draw=none,fill=steelblue31119180] (axis cs:6.7,0) rectangle (axis cs:6.9,1.08317);
\draw[draw=none,fill=steelblue31119180] (axis cs:7.7,0) rectangle (axis cs:7.9,0.6472);
\draw[draw=none,fill=steelblue31119180] (axis cs:8.7,0) rectangle (axis cs:8.9,1.34494);
\draw[draw=none,fill=steelblue31119180] (axis cs:9.7,0) rectangle (axis cs:9.9,0.72797);
\draw[draw=none,fill=steelblue31119180] (axis cs:10.7,0) rectangle (axis cs:10.9,1.60129);
\draw[draw=none,fill=steelblue31119180] (axis cs:11.7,0) rectangle (axis cs:11.9,0.85006);
\draw[draw=none,fill=steelblue31119180] (axis cs:12.7,0) rectangle (axis cs:12.9,1.05141);
\draw[draw=none,fill=steelblue31119180] (axis cs:13.7,0) rectangle (axis cs:13.9,0.63679);
\draw[draw=none,fill=steelblue31119180] (axis cs:14.7,0) rectangle (axis cs:14.9,1.09892);
\draw[draw=none,fill=steelblue31119180] (axis cs:15.7,0) rectangle (axis cs:15.9,0.6481);
\draw[draw=none,fill=darkorange25512714] (axis cs:-0.1,0) rectangle (axis cs:0.1,0.46044);
\addlegendimage{ybar,ybar legend,draw=none,fill=darkorange25512714}
\addlegendentry{Energy Focus}

\draw[draw=none,fill=darkorange25512714] (axis cs:0.9,0) rectangle (axis cs:1.1,0.51956);
\draw[draw=none,fill=darkorange25512714] (axis cs:1.9,0) rectangle (axis cs:2.1,1.08011);
\draw[draw=none,fill=darkorange25512714] (axis cs:2.9,0) rectangle (axis cs:3.1,1.17307);
\draw[draw=none,fill=darkorange25512714] (axis cs:3.9,0) rectangle (axis cs:4.1,1.32804);
\draw[draw=none,fill=darkorange25512714] (axis cs:4.9,0) rectangle (axis cs:5.1,1.07738);
\draw[draw=none,fill=darkorange25512714] (axis cs:5.9,0) rectangle (axis cs:6.1,1.09571);
\draw[draw=none,fill=darkorange25512714] (axis cs:6.9,0) rectangle (axis cs:7.1,0.50701);
\draw[draw=none,fill=darkorange25512714] (axis cs:7.9,0) rectangle (axis cs:8.1,0.56608);
\draw[draw=none,fill=darkorange25512714] (axis cs:8.9,0) rectangle (axis cs:9.1,0.56045);
\draw[draw=none,fill=darkorange25512714] (axis cs:9.9,0) rectangle (axis cs:10.1,0.61951);
\draw[draw=none,fill=darkorange25512714] (axis cs:10.9,0) rectangle (axis cs:11.1,0.65417);
\draw[draw=none,fill=darkorange25512714] (axis cs:11.9,0) rectangle (axis cs:12.1,0.7021);
\draw[draw=none,fill=darkorange25512714] (axis cs:12.9,0) rectangle (axis cs:13.1,0.50241);
\draw[draw=none,fill=darkorange25512714] (axis cs:13.9,0) rectangle (axis cs:14.1,0.54836);
\draw[draw=none,fill=darkorange25512714] (axis cs:14.9,0) rectangle (axis cs:15.1,0.51195);
\draw[draw=none,fill=darkorange25512714] (axis cs:15.9,0) rectangle (axis cs:16.1,0.55922);
\draw[draw=none,fill=forestgreen4416044] (axis cs:0.1,0) rectangle (axis cs:0.3,1.68125);
\addlegendimage{ybar,ybar legend,draw=none,fill=forestgreen4416044}
\addlegendentry{Runtime Focus}

\draw[draw=none,fill=forestgreen4416044] (axis cs:1.1,0) rectangle (axis cs:1.3,0.67671);
\draw[draw=none,fill=forestgreen4416044] (axis cs:2.1,0) rectangle (axis cs:2.3,1.08836);
\draw[draw=none,fill=forestgreen4416044] (axis cs:3.1,0) rectangle (axis cs:3.3,1.1765);
\draw[draw=none,fill=forestgreen4416044] (axis cs:4.1,0) rectangle (axis cs:4.3,1.33864);
\draw[draw=none,fill=forestgreen4416044] (axis cs:5.1,0) rectangle (axis cs:5.3,1.08465);
\draw[draw=none,fill=forestgreen4416044] (axis cs:6.1,0) rectangle (axis cs:6.3,1.10391);
\draw[draw=none,fill=forestgreen4416044] (axis cs:7.1,0) rectangle (axis cs:7.3,1.65934);
\draw[draw=none,fill=forestgreen4416044] (axis cs:8.1,0) rectangle (axis cs:8.3,0.72831);
\draw[draw=none,fill=forestgreen4416044] (axis cs:9.1,0) rectangle (axis cs:9.3,2.12944);
\draw[draw=none,fill=forestgreen4416044] (axis cs:10.1,0) rectangle (axis cs:10.3,0.83643);
\draw[draw=none,fill=forestgreen4416044] (axis cs:11.1,0) rectangle (axis cs:11.3,2.54842);
\draw[draw=none,fill=forestgreen4416044] (axis cs:12.1,0) rectangle (axis cs:12.3,0.99802);
\draw[draw=none,fill=forestgreen4416044] (axis cs:13.1,0) rectangle (axis cs:13.3,1.60041);
\draw[draw=none,fill=forestgreen4416044] (axis cs:14.1,0) rectangle (axis cs:14.3,0.72522);
\draw[draw=none,fill=forestgreen4416044] (axis cs:15.1,0) rectangle (axis cs:15.3,1.6859);
\draw[draw=none,fill=forestgreen4416044] (axis cs:16.1,0) rectangle (axis cs:16.3,0.73699);
\end{axis}

\end{tikzpicture}}
    \caption{GPT2-Large methods}
    \label{fig:gpt2largeopt}
\end{figure}
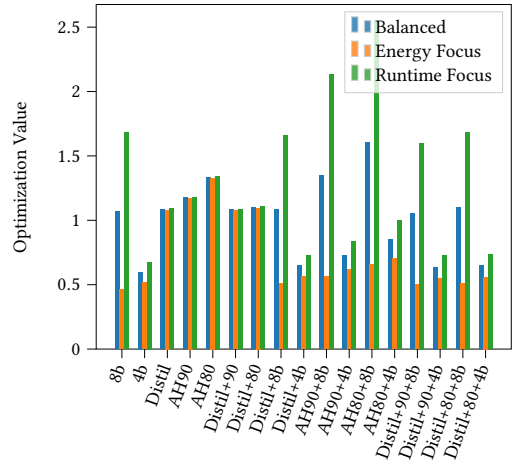

The custom optimization equation proved useful in balancing these metrics based on specific deployment goals, such as prioritizing energy savings or reducing runtime. Quantization was ideal for purely energy-focused optimization, whereas Knowledge Distillation balanced reductions across energy and time.

\subsubsection{Performance by Model Size}
The effectiveness of optimization techniques such as Quantization and Knowledge Distillation varies notably across different model sizes within the GPT-2 series (Regular, Large, and XL). These variations impact energy savings, runtime efficiency, and perplexity differently depending on the model size, as outlined below.

\paragraph{Quantization}
For smaller models like GPT-2 (125M), 8-bit quantization achieves high energy savings but results in significant runtime increases, with computation time tripling on average. In contrast, 4-bit quantization in these models offers a balanced trade-off, slightly increasing perplexity but maintaining relatively low runtime increases. This time increase is a known factor in the quantization method we use, \emph{BitsandBytes}; The developers inform that the method is intended primarily for model training, and is therefore not optimized for inference.

In larger models, such as GPT-2 Large (774M) and GPT-2 XL (1.5B), the relative impact of 8-bit quantization on runtime becomes more manageable, making it a viable option for scenarios where energy savings are prioritized. Notably, 4-bit quantization begins to deliver runtime reductions across these larger models, making it particularly effective for high-demand scenarios requiring both runtime and energy efficiency.

\paragraph{Knowledge Distillation}
The Knowledge Distillation (KD) method demonstrates consistent energy and runtime savings across model sizes due to reduced model size. However, we note that in the GPT-2 Large model, KD resulted in a relatively higher perplexity increase compared to the regular-sized DistilGPT-2 model. This is likely a result of differing model size ratios; Our distilled GPT2-Large model reduces the model size to 54\% of its teacher, while DistilGPT2 reduces the model size to 70\% of GPT2. The higher reduction in model size explains the higher increase in perplexity, as well as the greater reduction in time and energy. Results from other research, such as MiniLLM\cite{gu2024minillmknowledgedistillationlarge}, suggest that with sufficient training, increasing the size of a student model will allow it to retain more of a teacher model's instruction.

These insights indicate that both Quantization and Knowledge Distillation can be strategically applied based on model size and specific deployment priorities. Smaller models benefit more from 4-bit quantization for balanced efficiency, whereas larger models can leverage 8-bit quantization and KD for higher energy efficiency with minimal runtime trade-offs.
\paragraph{Attention Head Pruning}
Though Attention Head Pruning is notably less effective than other methods in our GPT-2 experiments, we observe that the method's perplexity increase becomes less pronounced in larger models, while the positive effects on speed and energy usage are amplified. This suggests that, as models become larger, a larger proportion of their attention heads become superfluous \& safely pruned.

\begin{table*}[h!]
\centering
\footnotesize
\begin{tabular}{|l|c|c|c|c|c|c|}
\hline
\textbf{Technique}       & \textbf{Perplexity} & \textbf{Perplexity Change (\%)} & \textbf{Runtime (s)} & \textbf{Runtime Change (\%)} & \textbf{Energy (kWh)} & \textbf{Energy Change (\%)} \\ \hline
Base                     & 26.93               & -                               & 1069.33              & -                             & 0.12282               & -                            \\ \hline
8-bit Quantization       & 26.92               & -0.04\%                         & 1962.33              & +83.59\%                      & 0.03784               & -69.17\%                     \\ \hline
4-bit Quantization       & 27.31               & +1.46\%                         & 729                  & -31.89\%                      & 0.06011               & -50.80\%                     \\ \hline
Knowledge Distillation   & 43.39               & +59.68\%                        & 569.67               & -46.72\%                      & 0.06481               & -47.22\%                     \\ \hline
AH90 (GPT2-L) & 31.83028 & +18.36\% & 979.3333 & -8.57\%  & 0.112075 & -8.93\%  \\ \hline
AH80 (GPT2-L) & 36.08654 & +34.27\% & 923.6667 & -13.71\% & 0.105043 & -14.58\% \\ \hline
\end{tabular}
\caption{Results for Standalone Techniques on GPT-2 Large (774M)}
\label{tab:gpt2_large_results}
\end{table*}

\begin{table*}[h!]
\centering
\footnotesize
\begin{tabular}{|l|c|c|c|c|c|c|}
\hline
\textbf{Technique}       & \textbf{Perplexity} & \textbf{Perplexity Change (\%)} & \textbf{Runtime (s)} & \textbf{Runtime Change (\%)} & \textbf{Energy (kWh)} & \textbf{Energy Change (\%)} \\ \hline
Base               & 17.67 &        & 655   &         & 0.066911 &         \\ \hline
8-bit Quantization & 17.67 & -0.05\% & 759.5 & 19.05\%  & 0.01897  & -71.51\% \\ \hline
4-bit Quantization & 19.1  & +8.14\%  & 355.5 & -44.37\% & 0.031103 & -53.24\% \\ \hline
AH90               & 20.04 & +14.16\% & 468.5 & -26.69\% & 0.054238 & -18.53\% \\ \hline
AH80               & 22.26 & +27.12\% & 439   & -31.30\% & 0.050512 & -24.10\% \\ \hline
\end{tabular}
\caption{Results for Standalone Techniques on GPT-2 XL (1.5B)}
\label{tab:gpt2_xl_results}
\end{table*}

Tables \ref{tab:gpt2_large_results} and \ref{tab:gpt2_xl_results} show the impact of each technique on larger models.

\subsection{Performance on Advanced Models and Hybrid Techniques}
As the capabilities of the experimental and hybrid models are far improved from the GPT-2 models previously analysed, we implement a selection of more specialized tests using the LM Evaluation Harness \cite{eval-harness}, including the following language \& knowledge tests:
\begin{itemize}
    \item \textbf{Arc\_challenge:} \cite{clark2018thinksolvedquestionanswering} A dataset composed of multiple-choice science questions, with a focus on problems requiring reasoning and in-depth scientific knowledge.

    \item \textbf{Hellaswag:} \cite{zellers2019hellaswagmachinereallyfinish} A commonsense reasoning test which evaluates a model’s ability to predict the most plausible continuation of a given situation, formulated as a multiple choice question. Questions are deliberately written to be confusing for LLMs, while trivial for humans.

    \item \textbf{MMLU (Massive Multitask Language Understanding):} \cite{hendrycks2021measuringmassivemultitasklanguage} A comprehensive multiple-choice benchmark made to test a model's general knowledge. Questions cover a wide degree of domains, including history, law, and various STEM fields.

    \item \textbf{TruthfulQA:} \cite{lin2022truthfulqameasuringmodelsmimic} A dataset of questions designed to evoke incorrect answers from LLMs, based on common misconceptions held by humans, \& sometimes inherited by LLMs. We utilize their MC2 task; A question is provided alongside a set of viable answers, potentially including multiple true and multiple false options. A model's score is the cumulative probability it assigns to the true answers.

    \item \textbf{Winogrande:} \cite{sakaguchi2019winograndeadversarialwinogradschema} A benchmark in which models, provided a sentence with two entities and an ambiguous pronoun, must decide which entity the pronoun refers to. Each sentence in the dataset comes in a pair, with one changed word which changes the final answer. This evaluates a model's ability to resolve ambiguities in language.
\end{itemize}
These benchmarks are chosen for their ability to evaluate distinct aspects of language comprehension and reasoning. For example, TruthfulQA tests factual reliability, while Hellaswag focuses on commonsense reasoning. Together, these tests provide a holistic assessment of a model's real-world applicability and robustness.
In addition, Perplexity is measured using the Wikitext-2\cite{merity2016pointer} dataset, and time/energy usage is observed using methodology identical to the previous set of tests. As quantization is not a measured factor in these tests, all models are loaded in bfloat16\cite{kalamkar2019studybfloat16deeplearning} format for consistency. By implementing this series of tests, we aim to capture deeper insights into the more specific knowledge and reasoning capabilities these larger models may be capable of, beyond those that can be depicted through perplexity alone.

To start our evaluation of the experimental models, we compare the MiniLLM-trained models to their KD, SeqKD, and from-scratch counterparts, across OPT and LLaMa frameworks. The results of these tests are shown in Tables \ref{tab:optmini} and \ref{tab:llamamini}. The best results of a given model type \& size are underlined; The best for a model using \emph{non-standard training} are listed in bold. The shot number for each logic test is written in brackets.

\begin{table*}[]
\begin{tabular}{|c|c|c|L|L|L|L|L|}
\hline
                                                  & Benchmark      & Metric    & OPT       & MiniLLM OPT (6.7B) & OPT (6.7B) & OPT 6.7B (KD) & OPT 6.7B (SeqKD) \\ \cline{2-8} 
                                                  & Model Size     &           & 13B       & 6.7B               & 6.7B       & 6.7B          & 6.7B             \\ \hline
\multicolumn{1}{|c|}{\multirow{5}{*}{Logic Test}} & arc\_challenge(25) & acc\_norm & 0.3951    & \uline{\textbf{0.407}}              & 0.3976     & 0.2841        & 0.2679           \\ \cline{2-8} 
\multicolumn{1}{|c|}{}                            & hellaswag(10)      & acc\_norm & 0.712     & \uline{\textbf{0.677}}              & 0.6616     & 0.4975        & 0.4775           \\ \cline{2-8} 
\multicolumn{1}{|c|}{}                            & MMLU(5)           & acc       & 0.2473    & 0.2399             & 0.2404     & 0.2494        & \uline{\textbf{0.2555}}           \\ \cline{2-8} 
\multicolumn{1}{|c|}{}                            & truthfulqa(0)     & mc2 (acc) & 0.3395    & 0.352              & 0.3628     & \uline{\textbf{0.4039}}        & 0.3956           \\ \cline{2-8} 
\multicolumn{1}{|c|}{}                            & winogrande(5)     & acc       & 0.6851    & \textbf{0.6346}             & \uline{0.6409}     & 0.618         & 0.6196           \\ \hline
Perplexity                                        & Wikitext       &           & 13.81     & 14.69              & 15.19      & \uline{\textbf{14.44}}         & 15.62            \\ \hline
Speed                                             &                & s         & 41        & 27                 & 27         & 27            & 27               \\ \hline
Energy                                            &                & kWh       & 0.006883 & 0.003793           & 0.003824  & 0.003789     & 0.003868      \\ \hline

\end{tabular}
\caption{Results of MiniLLM OPT models across logic tests \& Wikitext perplexity.}
\label{tab:optmini}
\end{table*}

\begin{table*}[]
\centering
\footnotesize
\begin{tabular}{|c|c|c|L|L|L|L|L|}
\hline
                                                  & Benchmark      & Metric    & Llama    & MiniLLM Llama (6.7B) & Llama (6.7B) & Llama 6.7B (KD) & Llama 6.7B (SeqKD) \\ \cline{2-8} 
                                                  & Model Size     &           & 13B      & 6.7B                 & 6.7B         & 6.7B            & 6.7B               \\ \hline
\multicolumn{1}{|c|}{\multirow{5}{*}{Logic Test}} & arc\_challenge(25) & acc\_norm & 0.555    & 0.524                & 0.530        & \uline{\textbf{0.538}}           & 0.524              \\ \cline{2-8} 
\multicolumn{1}{|c|}{}                            & hellaswag(10)      & acc\_norm & 0.812    & 0.780                & 0.788        & \uline{\textbf{0.803}}           & 0.798              \\ \cline{2-8} 
\multicolumn{1}{|c|}{}                            & MMLU(5)           & acc       & 0.464    & 0.343                & 0.349        & \uline{\textbf{0.360}}           & 0.352              \\ \cline{2-8} 
\multicolumn{1}{|c|}{}                            & truthfulqa(0)     & mc2 (acc) & 0.400    & 0.342                & 0.349        & 0.362           & \uline{\textbf{0.370}}              \\ \cline{2-8} 
\multicolumn{1}{|c|}{}                            & winogrande(5)     & acc       & 0.775    & \textbf{0.725}                & \uline{0.727}        & 0.707           & 0.704              \\ \hline
Perplexity                                        & Wikitext       &           & 6.36     & \textbf{7.5}                  & \uline{7.26}         & 9.09            & 9.28               \\ \hline
Speed                                             &                & s         & 50       & 31                   & 31           & 31              & 31                 \\ \hline
Energy                                            &                & kWh       & 0.008334 & 0.00485              & 0.004837     & 0.004824        & 0.00484            \\ \hline     
\end{tabular}
\caption{Results of MiniLLM Llama models across logic tests \& Wikitext perplexity.}
\label{tab:llamamini}
\end{table*}

To make sense of why models may perform better or worse than others, we identify the effects of each training method employed, and how they may affect a model's performance on each dataset. 
Models trained using MiniLLM tend to perform best on tasks requiring high-probability output prediction, like Arc\_challenge, Hellaswag, and Winogrande. By focusing on the highest probability outputs, ambiguous answers are chosen less often.
KD models often excel in TruthfulQA and perplexity tasks, benefiting from their broader distribution focus, which helps capture factual knowledge and language patterns. However, the GPT2 and OPT versions struggle with tests like Arc\_challenge and Hellaswag that require focused output selection.
SeqKD performs strongly on MMLU and occasionally on TruthfulQA, indicating that distilling sequence-level information can benefit tasks involving pure tests of knowledge, but may not always generalize to other test types.

\subsection{Further Insights from Modular and Hybrid Techniques}
This section explores the efficiency gains observed in models like ShearedLlama, MN-Minitron, and Llama-3.1-Minitron, which leverage hybrid techniques and unique methods to attain improved energy efficiency and speed. By combining advanced compression methods with optimized architectures, these models retain a majority of their performance even after compression.

\begin{table*}[h!]
\begin{tabular}{|c|c|c|L|L|L|L|L|L|L|}
\hline
                                                  & Benchmark      & Metric    & Llama2   6.7B & Sheared (2.7B) & Sheared (1.3B) & Mistral-Nemo Base & MN-Minitron & Llama3.1 & Llama-3.1-Minitron (W) \\ \cline{2-10} 
                                                  & Model Size     &           & 6.7B          & 2.7B           & 1.3B           & 12B               & 8B          & 8B       & 4B                     \\ \hline
\multicolumn{1}{|c|}{\multirow{5}{*}{Logic Test}} & arc\_challenge & acc\_norm & 0.523         & 0.430          & 0.298          & 0.651             & 0.644       & 0.579    & 0.556                  \\ \cline{2-10} 
\multicolumn{1}{|c|}{}                            & hellaswag      & acc\_norm & 0.790         & 0.710          & 0.610          & 0.852             & 0.830       & 0.816    & 0.761                  \\ \cline{2-10} 
\multicolumn{1}{|c|}{}                            & MMLU           & acc       & 0.458         & 0.265          & 0.252          & 0.690             & 0.695       & 0.653    & 0.605                  \\ \cline{2-10} 
\multicolumn{1}{|c|}{}                            & truthfulqa     & mc2 (acc) & 0.389         & 0.365          & 0.368          & 0.498             & 0.476       & 0.450    & 0.429                  \\ \cline{2-10} 
\multicolumn{1}{|c|}{}                            & winogrande     & acc       & 0.743         & 0.666          & 0.583          & 0.822             & 0.804       & 0.773    & 0.735                  \\ \hline
\multicolumn{1}{|l|}{Perplexity}                  & Wikitext       &           & 6.78          & 8.64           & 10.22          & 7.81              & 7.81        & 8.18     & 9.9                    \\ \hline
\multicolumn{1}{|l|}{Speed}                       &                & s         & 31            & 23             & 18             & 46                & 35          & 31       & 22                     \\ \hline
\multicolumn{1}{|l|}{Energy}                      &                & kWh       & 0.004859      & 0.002283       & 0.001301       & 0.007574          & 0.005186    & 0.004753 & 0.002844               \\ \hline
\end{tabular}
\caption{Results of additional models (Sheared-Llama, MN-Minitron, Llama-3.1-Minitron). Results marked * are sourced from respective model papers.}
\label{tab:advanced_models}
\end{table*}

Table \ref{tab:advanced_models} illustrates the energy and runtime benefits observed with the Sheared-Llama, MN-Minitron and Llama-3.1-Minitron models.
\subsubsection{ShearedLlama}
In our experiments on the ShearedLlama models, we find a significant decrease in accuracy on Arc\_challenge and MMLU, with a moderate score decrease on Winogrande. The pruning process likely removes a notable degree of specific knowledge, required for Arc\_challenge and MMLU. The removal of components may also harm the model's capability to handle complicated contexts, which affects its performance on Winogrande. Fortunately, the model retains reasonable performance on the more-generalized perplexity evaluation.
\subsubsection{Minitron Models}
For these models, we observe a significantly lower reduction in model capabilities. The Minitron method includes various techniques to ensure minimal loss in model knowledge; structured pruning, knowledge distillation, and the retention of knowledge from pruned heads all have a part in ensuring the preservation of the model's performance. The most notable observation is that MN-Minitron retains exactly the same level of perplexity as its teacher on Wikitext; On relevant tasks, the model would theoretically see no loss in performance.

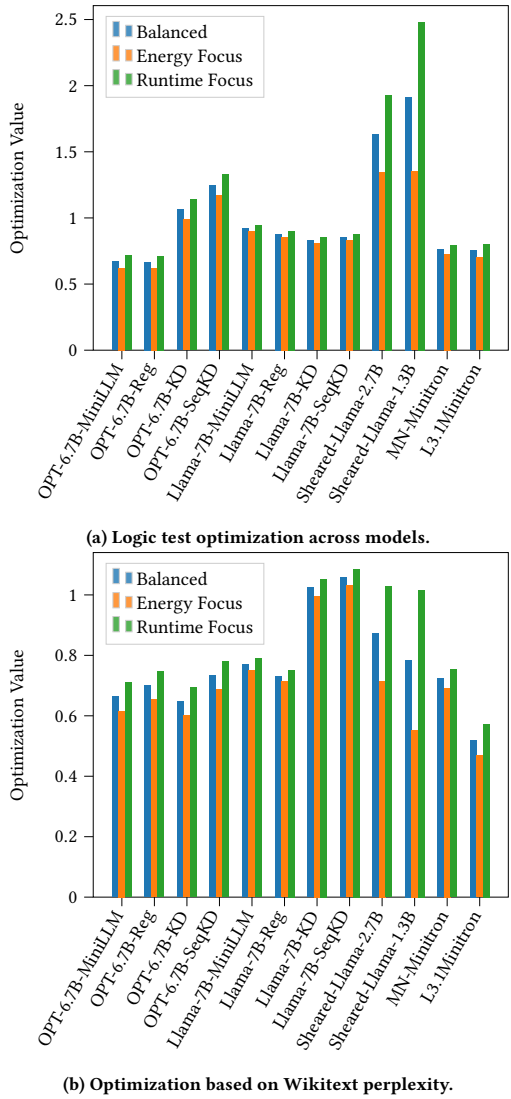
\begin{figure}[ht]
    \centering
    \begin{subfigure}[t]{0.8\columnwidth} 
        \centering
        \resizebox{\columnwidth}{!}{
\begin{tikzpicture}

\definecolor{darkgray176}{RGB}{176,176,176}
\definecolor{darkorange25512714}{RGB}{255,127,14}
\definecolor{forestgreen4416044}{RGB}{44,160,44}
\definecolor{lightgray204}{RGB}{204,204,204}
\definecolor{steelblue31119180}{RGB}{31,119,180}

\begin{axis}[
legend cell align={left},
legend style={
  fill opacity=0.8,
  draw opacity=1,
  text opacity=1,
  at={(0.03,0.97)},
  anchor=north west,
  draw=lightgray204
},
tick align=outside,
tick pos=left,
x grid style={darkgray176},
xmin=-0.88, xmax=11.88,
xtick style={color=black},
xtick={0,1,2,3,4,5,6,7,8,9,10,11},
xticklabel style={rotate=60.0,anchor=east},
xticklabels={
  OPT-6.7B-MiniLLM,
  OPT-6.7B-Reg,
  OPT-6.7B-KD,
  OPT-6.7B-SeqKD,
  Llama-7B-MiniLLM,
  Llama-7B-Reg,
  Llama-7B-KD,
  Llama-7B-SeqKD,
  Sheared-Llama-2.7B,
  Sheared-Llama-1.3B,
  MN-Minitron,
  L3.1Minitron
},
y grid style={darkgray176},
ylabel={Optimization Value},
ymin=0, ymax=2.6051347035,
ytick style={color=black}
]
\draw[draw=none,fill=steelblue31119180] (axis cs:-0.3,0) rectangle (axis cs:-0.1,0.66968452);
\addlegendimage{ybar,ybar legend,draw=none,fill=steelblue31119180}
\addlegendentry{Balanced}

\draw[draw=none,fill=steelblue31119180] (axis cs:0.7,0) rectangle (axis cs:0.9,0.667407177);
\draw[draw=none,fill=steelblue31119180] (axis cs:1.7,0) rectangle (axis cs:1.9,1.062630116);
\draw[draw=none,fill=steelblue31119180] (axis cs:2.7,0) rectangle (axis cs:2.9,1.250386388);
\draw[draw=none,fill=steelblue31119180] (axis cs:3.7,0) rectangle (axis cs:3.9,0.918925933);
\draw[draw=none,fill=steelblue31119180] (axis cs:4.7,0) rectangle (axis cs:4.9,0.878329185);
\draw[draw=none,fill=steelblue31119180] (axis cs:5.7,0) rectangle (axis cs:5.9,0.829440576);
\draw[draw=none,fill=steelblue31119180] (axis cs:6.7,0) rectangle (axis cs:6.9,0.854328414);
\draw[draw=none,fill=steelblue31119180] (axis cs:7.7,0) rectangle (axis cs:7.9,1.635458581);
\draw[draw=none,fill=steelblue31119180] (axis cs:8.7,0) rectangle (axis cs:8.9,1.915691408);
\draw[draw=none,fill=steelblue31119180] (axis cs:9.7,0) rectangle (axis cs:9.9,0.759066826);
\draw[draw=none,fill=steelblue31119180] (axis cs:10.7,0) rectangle (axis cs:10.9,0.751760443);
\draw[draw=none,fill=darkorange25512714] (axis cs:-0.1,0) rectangle (axis cs:0.1,0.622052641);
\addlegendimage{ybar,ybar legend,draw=none,fill=darkorange25512714}
\addlegendentry{Energy Focus}

\draw[draw=none,fill=darkorange25512714] (axis cs:0.9,0) rectangle (axis cs:1.1,0.622161927);
\draw[draw=none,fill=darkorange25512714] (axis cs:1.9,0) rectangle (axis cs:2.1,0.986617619);
\draw[draw=none,fill=darkorange25512714] (axis cs:2.9,0) rectangle (axis cs:3.1,1.171212124);
\draw[draw=none,fill=darkorange25512714] (axis cs:3.9,0) rectangle (axis cs:4.1,0.89564614);
\draw[draw=none,fill=darkorange25512714] (axis cs:4.9,0) rectangle (axis cs:5.1,0.855107767);
\draw[draw=none,fill=darkorange25512714] (axis cs:5.9,0) rectangle (axis cs:6.1,0.806677489);
\draw[draw=none,fill=darkorange25512714] (axis cs:6.9,0) rectangle (axis cs:7.1,0.831987541);
\draw[draw=none,fill=darkorange25512714] (axis cs:7.9,0) rectangle (axis cs:8.1,1.341776536);
\draw[draw=none,fill=darkorange25512714] (axis cs:8.9,0) rectangle (axis cs:9.1,1.350302146);
\draw[draw=none,fill=darkorange25512714] (axis cs:9.9,0) rectangle (axis cs:10.1,0.727073979);
\draw[draw=none,fill=darkorange25512714] (axis cs:10.9,0) rectangle (axis cs:11.1,0.700542717);
\draw[draw=none,fill=forestgreen4416044] (axis cs:0.1,0) rectangle (axis cs:0.3,0.717316399);
\addlegendimage{ybar,ybar legend,draw=none,fill=forestgreen4416044}
\addlegendentry{Runtime Focus}

\draw[draw=none,fill=forestgreen4416044] (axis cs:1.1,0) rectangle (axis cs:1.3,0.712652427);
\draw[draw=none,fill=forestgreen4416044] (axis cs:2.1,0) rectangle (axis cs:2.3,1.138642612);
\draw[draw=none,fill=forestgreen4416044] (axis cs:3.1,0) rectangle (axis cs:3.3,1.329560652);
\draw[draw=none,fill=forestgreen4416044] (axis cs:4.1,0) rectangle (axis cs:4.3,0.942205725);
\draw[draw=none,fill=forestgreen4416044] (axis cs:5.1,0) rectangle (axis cs:5.3,0.901550603);
\draw[draw=none,fill=forestgreen4416044] (axis cs:6.1,0) rectangle (axis cs:6.3,0.852203662);
\draw[draw=none,fill=forestgreen4416044] (axis cs:7.1,0) rectangle (axis cs:7.3,0.876669287);
\draw[draw=none,fill=forestgreen4416044] (axis cs:8.1,0) rectangle (axis cs:8.3,1.929140626);
\draw[draw=none,fill=forestgreen4416044] (axis cs:9.1,0) rectangle (axis cs:9.3,2.48108067);
\draw[draw=none,fill=forestgreen4416044] (axis cs:10.1,0) rectangle (axis cs:10.3,0.791059673);
\draw[draw=none,fill=forestgreen4416044] (axis cs:11.1,0) rectangle (axis cs:11.3,0.802978169);
\end{axis}

\end{tikzpicture}}
        \caption{Logic test optimization across models.}
        \label{fig:test2logic}
    \end{subfigure}%
    \hfill
    \begin{subfigure}[t]{0.8\columnwidth} 
        \centering
        \resizebox{\columnwidth}{!}{
\begin{tikzpicture}

\definecolor{darkgray176}{RGB}{176,176,176}
\definecolor{darkorange25512714}{RGB}{255,127,14}
\definecolor{forestgreen4416044}{RGB}{44,160,44}
\definecolor{lightgray204}{RGB}{204,204,204}
\definecolor{steelblue31119180}{RGB}{31,119,180}

\begin{axis}[
legend cell align={left},
legend style={
  fill opacity=0.8,
  draw opacity=1,
  text opacity=1,
  at={(0.03,0.97)},
  anchor=north west,
  draw=lightgray204
},
tick align=outside,
tick pos=left,
x grid style={darkgray176},
xmin=-0.88, xmax=11.88,
xtick style={color=black},
xtick={0,1,2,3,4,5,6,7,8,9,10,11},
xticklabel style={rotate=60.0,anchor=east},
xticklabels={
  OPT-6.7B-MiniLLM,
  OPT-6.7B-Reg,
  OPT-6.7B-KD,
  OPT-6.7B-SeqKD,
  Llama-7B-MiniLLM,
  Llama-7B-Reg,
  Llama-7B-KD,
  Llama-7B-SeqKD,
  Sheared-Llama-2.7B,
  Sheared-Llama-1.3B,
  MN-Minitron,
  L3.1Minitron
},
y grid style={darkgray176},
ylabel={Optimization Value},
ymin=0, ymax=1.14014263545,
ytick style={color=black}
]
\draw[draw=none,fill=steelblue31119180] (axis cs:-0.3,0) rectangle (axis cs:-0.1,0.663485094);
\addlegendimage{ybar,ybar legend,draw=none,fill=steelblue31119180}
\addlegendentry{Balanced}

\draw[draw=none,fill=steelblue31119180] (axis cs:0.7,0) rectangle (axis cs:0.9,0.700325771);
\draw[draw=none,fill=steelblue31119180] (axis cs:1.7,0) rectangle (axis cs:1.9,0.646318553);
\draw[draw=none,fill=steelblue31119180] (axis cs:2.7,0) rectangle (axis cs:2.9,0.734056012);
\draw[draw=none,fill=steelblue31119180] (axis cs:3.7,0) rectangle (axis cs:3.9,0.769588058);
\draw[draw=none,fill=steelblue31119180] (axis cs:4.7,0) rectangle (axis cs:4.9,0.731965422);
\draw[draw=none,fill=steelblue31119180] (axis cs:5.7,0) rectangle (axis cs:5.9,1.024243621);
\draw[draw=none,fill=steelblue31119180] (axis cs:6.7,0) rectangle (axis cs:6.9,1.058178532);
\draw[draw=none,fill=steelblue31119180] (axis cs:7.7,0) rectangle (axis cs:7.9,0.87165772);
\draw[draw=none,fill=steelblue31119180] (axis cs:8.7,0) rectangle (axis cs:8.9,0.78499085);
\draw[draw=none,fill=steelblue31119180] (axis cs:9.7,0) rectangle (axis cs:9.9,0.722789746);
\draw[draw=none,fill=steelblue31119180] (axis cs:10.7,0) rectangle (axis cs:10.9,0.520076688);
\draw[draw=none,fill=darkorange25512714] (axis cs:-0.1,0) rectangle (axis cs:0.1,0.616294155);
\addlegendimage{ybar,ybar legend,draw=none,fill=darkorange25512714}
\addlegendentry{Energy Focus}

\draw[draw=none,fill=darkorange25512714] (axis cs:0.9,0) rectangle (axis cs:1.1,0.652848884);
\draw[draw=none,fill=darkorange25512714] (axis cs:1.9,0) rectangle (axis cs:2.1,0.600085827);
\draw[draw=none,fill=darkorange25512714] (axis cs:2.9,0) rectangle (axis cs:3.1,0.687575704);
\draw[draw=none,fill=darkorange25512714] (axis cs:3.9,0) rectangle (axis cs:4.1,0.750091546);
\draw[draw=none,fill=darkorange25512714] (axis cs:4.9,0) rectangle (axis cs:5.1,0.712613595);
\draw[draw=none,fill=darkorange25512714] (axis cs:5.9,0) rectangle (axis cs:6.1,0.996134379);
\draw[draw=none,fill=darkorange25512714] (axis cs:6.9,0) rectangle (axis cs:7.1,1.030506934);
\draw[draw=none,fill=darkorange25512714] (axis cs:7.9,0) rectangle (axis cs:8.1,0.715132679);
\draw[draw=none,fill=darkorange25512714] (axis cs:8.9,0) rectangle (axis cs:9.1,0.553311888);
\draw[draw=none,fill=darkorange25512714] (axis cs:9.9,0) rectangle (axis cs:10.1,0.692325891);
\draw[draw=none,fill=darkorange25512714] (axis cs:10.9,0) rectangle (axis cs:11.1,0.467469824);
\draw[draw=none,fill=forestgreen4416044] (axis cs:0.1,0) rectangle (axis cs:0.3,0.710676033);
\addlegendimage{ybar,ybar legend,draw=none,fill=forestgreen4416044}
\addlegendentry{Runtime Focus}

\draw[draw=none,fill=forestgreen4416044] (axis cs:1.1,0) rectangle (axis cs:1.3,0.747802657);
\draw[draw=none,fill=forestgreen4416044] (axis cs:2.1,0) rectangle (axis cs:2.3,0.692551279);
\draw[draw=none,fill=forestgreen4416044] (axis cs:3.1,0) rectangle (axis cs:3.3,0.78053632);
\draw[draw=none,fill=forestgreen4416044] (axis cs:4.1,0) rectangle (axis cs:4.3,0.78908457);
\draw[draw=none,fill=forestgreen4416044] (axis cs:5.1,0) rectangle (axis cs:5.3,0.75131725);
\draw[draw=none,fill=forestgreen4416044] (axis cs:6.1,0) rectangle (axis cs:6.3,1.052352863);
\draw[draw=none,fill=forestgreen4416044] (axis cs:7.1,0) rectangle (axis cs:7.3,1.085850129);
\draw[draw=none,fill=forestgreen4416044] (axis cs:8.1,0) rectangle (axis cs:8.3,1.028182762);
\draw[draw=none,fill=forestgreen4416044] (axis cs:9.1,0) rectangle (axis cs:9.3,1.016669813);
\draw[draw=none,fill=forestgreen4416044] (axis cs:10.1,0) rectangle (axis cs:10.3,0.753253601);
\draw[draw=none,fill=forestgreen4416044] (axis cs:11.1,0) rectangle (axis cs:11.3,0.572683552);
\end{axis}

\end{tikzpicture}}
        \caption{Optimization based on Wikitext perplexity.}
        \label{fig:test2perp}
    \end{subfigure}
    \caption{Comparison of optimization across tested models on logic and perplexity metrics.}
    \label{fig:optimization_comparison}
\end{figure}

\section{Discussion}
This study confirms that modular and hybrid optimization techniques deliver meaningful gains in energy efficiency and processing speed across various large language model architectures. By organizing optimizations to prioritize different deployment needs—be it energy efficiency, computational speed, or a balanced configuration—model adaptations can be effectively tailored to meet specific operational requirements.

Results indicate that certain methods, such as 4-Bit Quantization, are particularly effective for energy-limited applications, while Knowledge Distillation supports balanced optimization, retaining a reasonable measure of accuracy while also reducing computational load. Hybrid methods, exemplified by the Minitron approach, achieved significant reductions in both energy usage and runtime with minimal accuracy loss, making them ideal for high-throughput, low-latency environments.

In the following, we provide a detailed comparative analysis of each optimization technique, evaluating their trade-offs and situating them within real-world deployment contexts.

\subsection{Comparative Analysis of Optimization Techniques}

Table \ref{tab:optimization_summary} summarizes the comparative performance of the tested optimization techniques across critical metrics, evaluating their effectiveness in energy reduction, computation time, and perplexity retention. The results emphasize the flexibility needed to prioritize specific operational goals based on model requirements and use cases.
The following insights further underscore how these techniques align with distinct deployment needs:
\begin{table}[h]
\centering
\scriptsize 
\begin{tabular}{|l|c|c|c|}
\hline
\textbf{Technique} & \textbf{Energy Savings} & \textbf{Time Reduction} & \textbf{Perplexity Change} \\ \hline
4-bit Quantization         & High         & Mod. Increase & Low     \\ \hline
Knowledge Distillation     & Moderate     & Moderate      & Moderate   \\ \hline
Attention Head Pruning     & Low          & Low           & High    \\ \hline
Magnitude Pruning          & Moderate     & Moderate           & High   \\ \hline
Hybrid Techniques          & High         & High          & Low       \\ \hline
\end{tabular}
\caption{Summary of Optimization Techniques and Their Resource-Performance Trade-offs}
\label{tab:optimization_summary}
\end{table}

4-bit Quantization consistently achieved the highest energy savings with only a modest increase in perplexity, making it suitable for applications where energy efficiency is paramount. Knowledge Distillation yielded moderate reductions in both energy and computation time with balanced perplexity retention, supporting applications where resource efficiency is necessary without heavily compromising accuracy. Conversely, Attention Head Pruning exhibited limited reductions in energy and time but led to noticeable increases in perplexity, particularly for generative tasks that require high accuracy. Hybrid techniques, such as those implemented in MN-Minitron, achieved significant reductions across both energy and time metrics with minimal loss in perplexity, indicating their suitability for scenarios where maximum efficiency is essential without sacrificing performance. In summary, when energy savings are prioritized, 4-bit Quantization and hybrid techniques are recommended for their efficiency with minimal performance trade-offs. In contrast, Knowledge Distillation provided balanced reductions across metrics, making them ideal for tasks with moderate accuracy and latency demands.

\subsection{Evaluating Trade-offs in Energy and Performance}

Each optimization method explored here reflects specific trade-offs among energy savings, computational speed, and perplexity retention. Quantization, particularly in 4-bit format, led to notable energy reductions with minimal perplexity increase, though it slightly increased computation time—making it particularly suitable for energy-sensitive tasks. Knowledge Distillation balanced reductions across both energy and time, with a moderate impact on perplexity, supporting deployments where both accuracy and resource efficiency are moderately prioritized. NVIDIA's Minitron method achieved high efficiency in both energy and time with minimal performance loss, making them ideal for any scenario.

\subsection{Interpretation of Optimization Equation Outcomes}

Our custom optimization equation provided a quantitative assessment of the trade-offs between energy, time, and perplexity, with adjustable weights that enabled prioritization of different metrics based on specific deployment needs. For example, when energy savings were prioritized, 4-bit Quantization and MN-Minitron achieved optimal scores due to their high energy efficiency. Conversely, when prioritizing computational speed, Knowledge Distillation performed better due to its substantial reduction in processing time. The equation also highlighted scenarios where slight increases in perplexity were warranted for significant gains in energy efficiency, as observed with Knowledge Distillation. Figures ~\ref{fig:gpt2opt}, ~\ref{fig:gpt2largeopt}, ~\ref{fig:gpt2xlopt}, and \ref{fig:optimization_comparison} visually represent these trade-offs, providing further insights into how each technique aligns with various operational requirements.

\subsection{Implications for Sustainable LLM Deployment}

Our findings underscore the potential for deploying optimized LLMs in resource-constrained environments. Notably, basic methods, combinations, and experimental methods all demonstrate that meaningful reductions in resource demands are achievable without substantial performance degradation, advancing the goal of sustainable LLM deployment. The flexibility provided by balanced, energy-focused, and runtime-focused configurations in the optimization equation enables customization based on computational and operational requirements. 

These results support the feasibility of deploying efficient LLMs across a range of hardware infrastructures, from cloud-based GPUs to mobile devices with limited computational power. This adaptability is especially valuable in industries like healthcare, finance, and mobile applications, where sustainable and accessible AI is crucial.

\section{Conclusions and Future Work}

This study systematically evaluated various optimization strategies for LLMs, focusing on achieving a balance between resource efficiency and model performance. Our results highlighted that while standalone techniques can be effective for targeted improvements, hybrid approaches such as Knowledge Distillation combined with Quantization or Pruning provide the most consistent trade-offs across all metrics.

When optimization techniques were combined, the pairing of 4-Bit Quantization with Knowledge Distillation yielded the best balance between efficiency and performance retention. Advanced models like MN-Minitron and Llama-3.1-Minitron showcased the advantages of structured pruning and knowledge retention strategies, especially in high-throughput applications. For instance, Llama-3.1-Minitron achieved a 50\% reduction in model size with minimal accuracy loss, while MN-Minitron achieved a 33\% size reduction while retaining high accuracy, reinforcing the utility of hybrid optimization approaches for specific industrial applications.

Looking ahead, future work will involve refining the optimization framework, and extending the testing procedures for future methods, architectures, and benchmarks. In order to allow for an adjustable and broadly applicable methodology, the optimization equation is relatively simple; more in-depth calculations could come closer to our intended balance, and include additional factors such as energy costs for training. It will also help to investigate cost-effective retraining strategies to counteract perplexity increases from compression, such as those employed by NVIDIA for Minitron. As we see from the ShearedLlama experiments, even newer methods can cause performance degradation; finding ways to counteract these losses could allow for a wide degree of alternative methods to become viable.
In addition, we can place focus on environmental impacts by creating models optimized for specific tasks and deployment constraints, such as low-power or latency-sensitive environments. It will also be imperative for research to be conducted into refining the fundamental methods; Although Quantization is powerful, the time concerns seen in popular implementations of the method may turn some users away. By advocating for future model paradigms to be designed with quantization and other methods in mind, it could be possible to eliminate any negative impacts and increase the energy savings.

Another potential avenue for investigation is the training costs incurred by the methods, where applicable. Resulting from the surging popularity of large-scale services like ChatGPT or Google Gemini, seeing several million users every day, most of our focus is placed on costs of inference. However, smaller-scale operations are likely to use much more energy on training a model rather than running it. As such, future research into this topic should make considerations regarding the costs of particular training methods, ensuring all aspects of a model's training, compression, and usage can be made as efficient as possible. 

These advancements would further support the goal of sustainable, scalable, and accessible AI across industries.

\pagebreak

\bibliographystyle{plain}
\bibliography{arxivsubmit}
\end{document}